# Human machine interaction systems encounters convergence

Josephine Selvarani Ruth D and Vishwas Navada B

*Abstract*—Human machine interaction systems are those of much needed in the emerging technology to make the user aware of what is happening around. It is huge domain in which the smart material enables the factor of convergence. One such is the piezoelectric crystals, is a class of smart material and this has an incredible property of self-sensing actuation (SSA). This property of SSA has added an indescribable advantage to the robotic field by having the advantages of exhibiting both the functionality of sensing and actuating characteristics with reduced devices, space and power. This paper focuses on integrating the SSA to drive an unmanned ground vehicle with wireless radio control system which will be of great use in all the automation field. The piezo electric plate will be used as an input device to send the signal to move the UGV in certain direction and then, the same piezo-electric plate will be used as an actuator for haptic feedback with the help of drive circuit if obstacles or danger is experienced by UGV.

## I. OVERVIEW OF THE WORK

This article throws light on self-sensing actuation for various control applications. In closed-loop micromanipulation tasks, various sensors have been used in order to perform accurately with a fast response time. Unfortunately, these sensors are not ideally adapted to the micro and nano world because of their sizes, performances, and limited measurement of degrees of freedom. Hence, an alternative method is to use the actuators in sensing mode by capturing the variation in electrical resistance on its actuation by the concept of self-sensing actuation method. There are several advantages of the self-sensing method compared to the use of external sensors. One of them is that it allows a consistent reduction in costs by eliminating expensive sensors. As can be seen, resolution can also be sub-micrometric and comparable to that of external sensors. Self-sensing is based on energy extraction in the form of electrical variable; in fact, displacement is nearly proportional to the output voltage i.e., from a mechanical varying component, the appropriate exact variation in the electrical component of an actuating element is produced. The concept of combining sensing and actuating features in the same device leads to the so-called self-sensing technique. The concept of combining sensing and actuating features in the same device was first started by the work of Dosch *et al.* (1992) in piezoelectric cantilever. Fig.1 illustrates the effects of a self-sensing approach in a general control application.

Self-sensing is applicable not only with piezoelectric materials or SMA but in general with materials that intrinsically contain information about mechanical quantities such as force and displacement as well as electrical quantities, which, in case of piezoelectric materials are charge and electric field and in case of SMA is electrical resistance due to strain during actuation. Hence in a self-sensing based control loop, the mechanical quantities are reconstructed through the measurement of electrical quantities and the model of the transducer. So from an instrumentation perspective, the final control element (actuator) and the transducer for feedback control (sensor) are of the same element. Hence, feedback control for a self-sensing control application works without the need of any sensor, since it is reconstruction based on the knowledge of model of the self-sensing actuator.

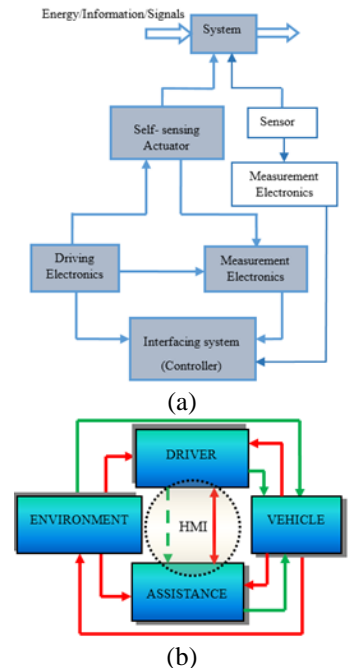

Fig. 1 Schematics of control
(a) Self-sensed actuators (b) Piezoelectric assistance in loop

The proposed conceptual technique is implemented by a piezoelectric based 'Piezaptic' (**PIEZ**oelectricity +h**APTIC**s) device integrated with the handheld remote control device that acts as a human machine interaction system. The device is instrumented with a pair of

*Research supported by DST INSPIRE & RBCCPS, IISc, Bangalore.

Josephine S Ruth D, INSPIRE Faculty is with the Robert Bosch Centre for Cyber Physical Systems, Indian Institute of Science, Bengaluru -560 012 (djsruth@gmail.com, rhosephine@iisc.ac.in ).

Vishwas Navada B, Project Assistant with the Robert Bosch Centre for Cyber Physical Systems, Indian Institute of Science, Bengaluru -560 012. (nvishwas@iisc.ac.in)

piezocrystals organized in a synergistic configuration which perform shared sensing and actuation, making it bi-functional; it functions as an active sensor for positioning the vehicle (pedal position sensing) to control the speed and as a haptic actuator to generate active vibration (haptic) feedback to the driver. The design is a proof-of-concept. The uniqueness of this work is the design of piezo based bi-functionality (two-in-one) module for the purpose of servoing and to create haptics in the handheld device; in addition, it is capable to take control input to servo the UGV. Similar work has been carried out using shape memory alloy wires to exhibit its bi-functionality in which the variation in the electrical resistance of the wires acts as the sensing signal and the haptics is deployed by its shape memory effect behavior Josephine.et al (2016).

This work is devoted to develop the principles and techniques needed to realize piezoelectric plates capable of haptic interaction. The paper starts with the design and construction of piezo based haptic master system which is the handheld device, its characteristics as a master and a haptic system. It is then used in a closed loop control and to manipulate slave robot i.e, the unmanned ground vehicle. The experimental results on this system for its motion and force feedback is tested and proved the piezo bi-functional capability in such a human machine interaction systems.

## II. DESIGN AND WORKING OF THE SENSAPTIC MASTER – SLAVE SYSTEMATH

Bilateral control is one of the control techniques to transmit kinesthetic sensation bi-directionally. In the proposed bilateral tele-operator, there are primarily two design goals which ensure a close coupling between the human operator and the vehicle in the remote environment. The first goal is that the slave manipulator should track the position of the master manipulator, and the second goal is that the environmental force acting on the slave when it contacts an article / detects impact in the remote environment, be accurately transmitted to the master.

In this haptic master – slave system, the piezo haptic master plays two roles: firstly it acts as the reference input device to the slave and secondly as a vibration display device from the slave environment to the operator. A force feedback from slave to the master representing contact information provides a more extensive sense of tele-presence. The uniqueness in this system is that a single configuration of piezo patches plays the dual roles: to sense the operator movements (in place of traditionally used sensors like encoder, potentiometer or accelerometer) and concurrently generates force for haptic feel (in place of conventionally used motors, etc.).

### A. Piezo sensaptic (SENSe+hAPTIC) master: Piezo for shared sensing and haptics

The master system is a handheld device with piezo plates arranges in a fashion by which the pressure i.e., an impulse input is taken as a input signal to control the vehicle and another piezo acts as an actuator which works on inverse piezoelectricity to generate vibration depending on the intensity of threat in the slave environment. The handheld device has unidirectional control over the slave for which the manipulator (1-DOF joystick) can be operated by the human to move about one axis bi-directionally, which controls the slave. Unlike the usual master which is structured with separate sensor and actuator, the piezo based sensaptic device that in unison performs both these functions is used for the development of the haptic master. This master system acts as a HMI where the force stimuli can provide the operator with interactive information (the intensity of threat) of the dynamic slave site as in fig.2.

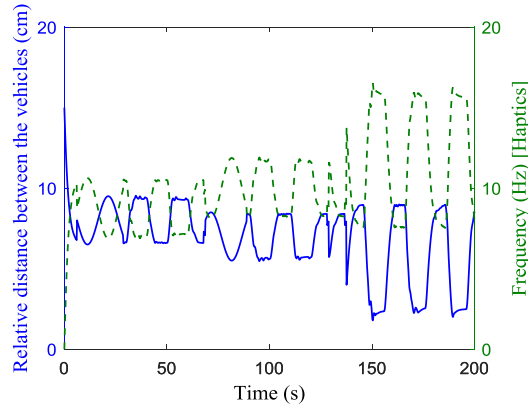

Fig.2 Input-output characteristic of Piezo haptic master in sensing mode

### B. Piezaptic master – slave in control loop

In the proposed piezo device based haptic master system, haptic interface is established such that it can exchange mechanical energy with a user as shown in Fig.3.

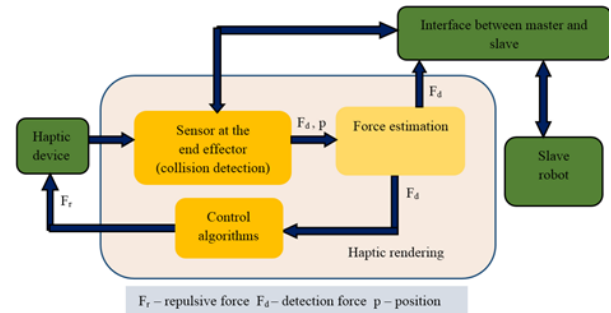

Fig. 3 Haptic rendering

The haptic body interface highlights the physical connection between operator and master system through which energy is exchanged. Haptic devices are input-output devices, meaning that they track the human's physical manipulations (input) and provide realistic force feedback sensations coordinated with on-site slave environment (output). Haptic interfaces are relatively sophisticated devices. When a human manipulates the handle of a haptic master device, the voltage developed in the active sensor (sensaptic piezo) is transmitted to an interface controller at very high rates. Here the information is processed to determine the position of the end effector of the slave.

If the end effector is obstructed by any object in its environment, its sensor sends the signal to the interface

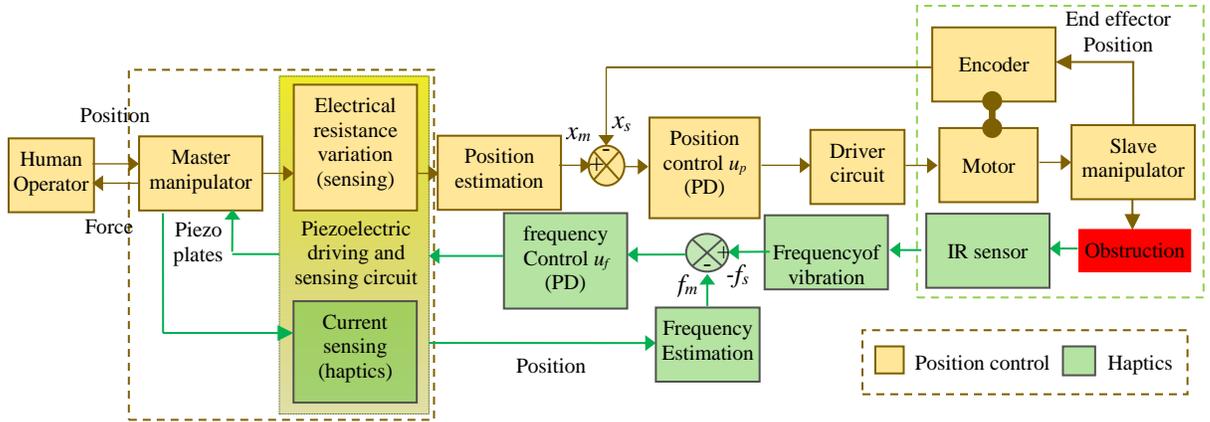

Fig. 5 Bilateral control in human - robot interaction system

Controller in the host computer. If the kinematics determines that a reaction force is required, the interface controller feeds back the repulsive force to the master device. Actuators (sensaptic Piezo) apply these forces by correlating the force experienced, which in turn simulates the desired sensations to the operator.

Thus the piezo sensaptic master does perfect servoing of the slave and also displays the force information to the operator with stable dynamic co-ordination of master – slave system.

*C. Implementation methodology*

Sensing and actuation are two major parts of a closed loop control system. In this application, we are demonstrating that the same senor can be used as an actuator i.e. piezo-electric plate. This will enable us to reduce the number of parts in the hardware unit. We demonstrate this with the help of a UGV.

Vibration or impulse of force on piezo electric material produces a spike of voltage pulse which is considered as input to the control system. This input is read by a microcontroller which sends this signal wirelessly to the UGV, based on the received signal UGV does the actuation part by moving motors. When the sensors on the UGV sense obstacles using its obstacle sensor it sends back the signal to the wireless controller. This signal is taken as input to haptic feedback system which basically alerts the person with the help of physical vibration. In this system piezo driver circuit drives piezo-electric material to generate a vibration.

Unmanned Ground Vehicle as name itself says there will be no person on board of vehicle to control it. It will be either autonomous i.e. which can decide its path based on input from external sensors such as camera and GPS, or it will be wirelessly controlled using a wireless transmitter and receiver units. Here we demonstrate experiment with the help of wirelessly controlled UGV. It consists of ATMEGA328 microcontroller as brain and HC SR-04 ultrasonic sensor for obstacle detection. Motors and motor drivers for movement.

## III. EXPERIMENTAL IMPLEMENTATION AND ANALYSIS

The bi-functional sensaptic master controls an unmanned ground vehicle. To initialize the system, the first step is to set the position of the motor to the zero position. The tacking performance of the senspatic master – slave system is practically determined using bilateral control system. An operator manipulates the master system for driving command of the slave robot and the corresponding responses were perfectly the same values as the corresponding inputs as presented in Fig. 6 (a). Each position response between the master and slave was corresponded i.e., the motors of the links in the slave robot take respective position, in order that the end effector is positioned according to the sensing signal commanded/manipulated by the master/operator. In Fig. 6 (b), the displacement of the slave robot is determined. The PD controller output to the driving circuit of the motor for the corresponding control action is shown in Fig. 6 (c). Figure 6 (d) displays the change in the displacement of the vehicle in the slave environment. When the slave encounters an object in the forward direction; the corresponding change in the ultrasonic sensor in the vehicle captures the intensity of the threat by the reduction in the distance which is sent as a control input to evoke the haptics feedback to the master device are shown in Fig 6. (e).

The response time of the first generation devices like electromagnetic actuators is about 35-60ms. The second generation devices using the smart materials like electro active polymers, piezoelectric crystals reduced the response time to 5-15 ms. The proposed SMA based sensaptic master performs stable operation with a response time of 0.930 ms.

## IV. CONCLUSION

A concept of shared sensing and actuation of piezo based device is demonstrated by a suitable/unique design that uses 2 piezos to individually perform its own function (sensing or actuation) alone within one interlinked flexible structure. The new technique is implemented by an integrated design of sensor and actuator, to display its independent bi-functionality as an active sensor and high frequency actuator by haptics.

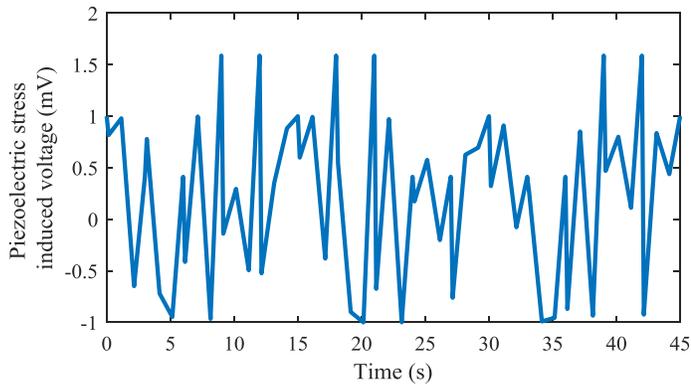

(a)

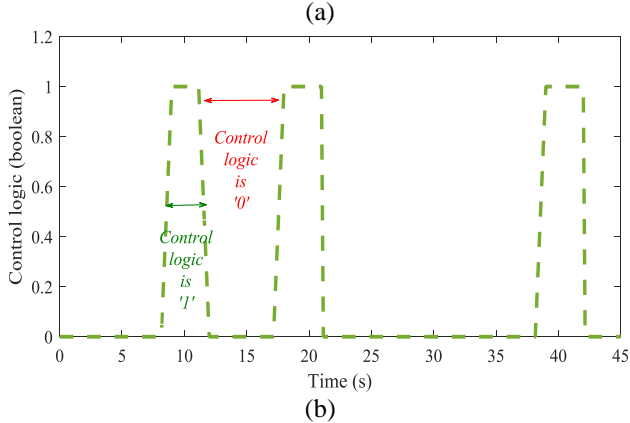

(b)

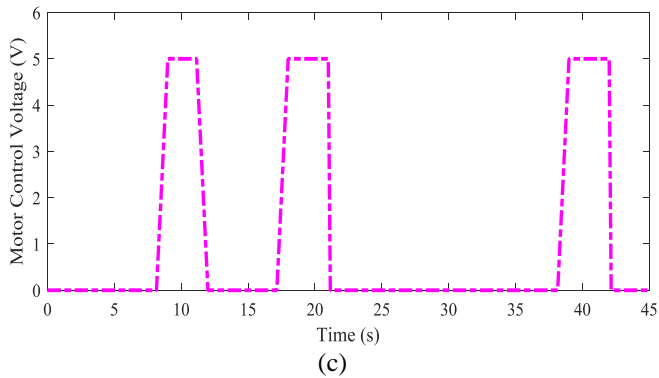

(c)

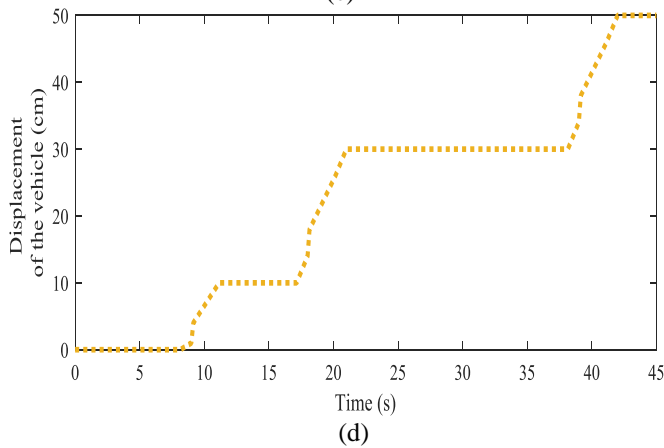

(d)

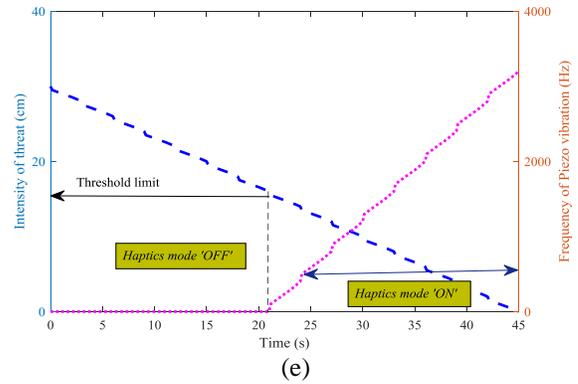

(e)

Fig.6. Tracking performance of senspatic master – slave system (a) position tracking (b) control logic tracking (c) controller output /motor control input (d) displacement of the slave robot (e) Frequency response of the haptics w.r.t intensity of the threat

ACKNOWLEDGMENT

The authors thank Roshni K S, Intern student at Robert Bosch Centre for Cyber Physical Systems, for her assistance in this work.

This work was financially supported in part by the Government of India Ministry of Science & Technology, Department of Science & Technology under grant no. DST/INSPIRE/04/2017/000533. The work is hosted at Robert Bosch Centre for Cyber Physical Systems, Indian Institute of Science, Bengaluru -560 012.